\pdfoutput=1

\documentclass[11pt]{article}

\PassOptionsToPackage{table,xcdraw}{xcolor}

\usepackage[preprint]{acl}

\usepackage{times}
\usepackage{latexsym}

\usepackage[T1]{fontenc}

\usepackage[utf8]{inputenc}

\usepackage{microtype}

\usepackage{inconsolata}

\usepackage{graphicx}
\usepackage{amsmath}
\usepackage{longtable}
\usepackage{multirow} 
\usepackage{xcolor} 
\usepackage{makecell}
\usepackage{hyperref}
\usepackage[capitalize,noabbrev]{cleveref}

\title{Encode Errors: Representational Retrieval of In-Context Demonstrations for Multilingual Grammatical Error Correction}

\author{Guangyue Peng, Wei Li, Wen Luo, Houfeng Wang\thanks{Corresponding author} \\
  State Key Laboratory of Multimedia Information Processing, \\ School of Computer Science, Peking University \\
  \texttt{\{agy,wanghf\}@pku.edu.cn}\\ 
  {\tt weili22@stu.pku.edu.cn,llvvvv22222@gmail.com} \\
  }

\begin{document}
\maketitle
\begin{abstract}
Grammatical Error Correction (GEC) involves detecting and correcting the wrong usage of grammar. While large language models (LLMs) with in-context learning (ICL) capabilities have shown significant progress on various natural language processing (NLP) tasks, their few-shot performance on GEC remains suboptimal. This is mainly due to the challenge of retrieving suitable in-context demonstrations that capture error patterns instead of semantic similarity. In this paper, we demonstrate that LLMs can inherently capture information related to grammatical errors through their internal states. From these states, we extract the Grammatical Error Representation (GER), an informative and semantically neutral encoding of grammatical errors. Our novel GER-based retrieval method significantly boosts performance in ICL settings on multilingual GEC datasets, improving the precision of correction. For high-resource languages, our results on 8B-sized open-source models match those of closed-source models such as Deepseek2.5 and GPT-4o-mini. For low-resource languages, our $F_{0.5}$ scores surpass the baseline by up to a factor of 1.20. This method provides a more precise and resource-efficient solution for multilingual GEC, offering a promising direction for interpretable GEC research.\footnote{Code is publicly available at \url{https://github.com/viniferagy/GER}.}
\end{abstract}

\section{Introduction}

\begin{figure}[t]
  \includegraphics[width=\columnwidth]{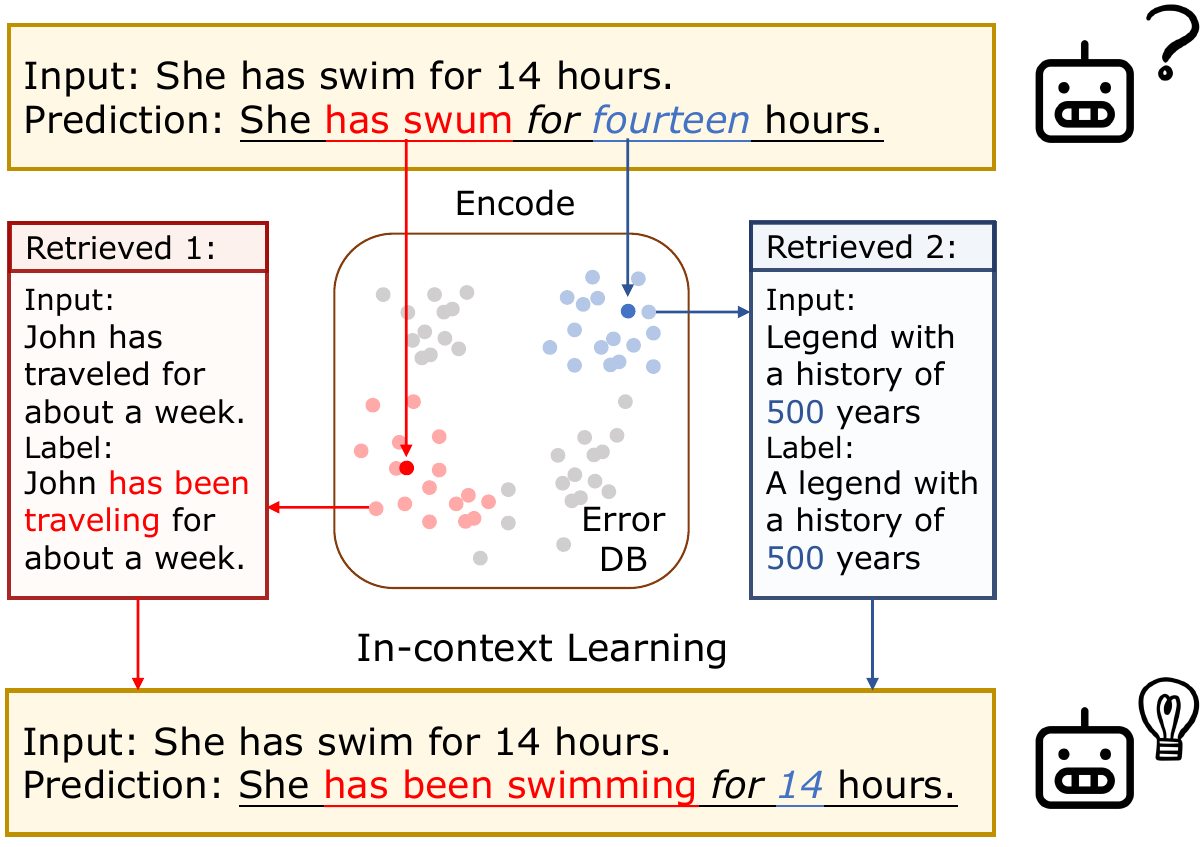}
  \caption {A minimal working example demonstrating the workflow of representational retrieval. Given an erroneous input with predictions containing both under-correction (marked in \textcolor{red}{red}) and over-correction (marked in \textcolor[rgb]{0.0,0.0,0.71}{blue}), we first transform the error information detected by the model into the Grammatical Error Representation (GER). Then, we retrieve GER-adjacent demonstrations from the error database, which exhibit error patterns similar to those in the input. These demonstrations guide the model to make more precise corrections and alleviate over-corrections.}
  \label{fig:intro}
\end{figure}

Grammatical Error Correction (GEC) is an important research field in natural language processing (NLP), as it requires language models to understand the syntax, semantics, and pragmatics underlying the subtle structures of natural sentences \citep{bryant2023grammatical}. Initially considered a specific case of machine translation \citep{yuan2016grammatical, junczys-dowmunt2018approaching}, GEC has evolved with two dominant approaches. Text-to-text methods \citep{katsumata2020stronger, sun2020instantaneous, ingolfsdottir2023byte} construct pairs of erroneous input and corrected output sentences and train encoder-decoder models, while text-to-edit approaches \citep{stahlberg2020seq, omelianchuk2020gector} rely on the encoder's capabilities to identify errors and make corrections. 

As Large Language Models (LLMs) come to prominence, they have achieved considerable results in GEC \citep{ maeng2023effectiveness, zeng2024evaluating}. However, LLMs that are not specifically adapted for GEC tasks face two main challenges: misalignment and over-correction \citep{loem2023exploring}. These models often produce corrections misaligned with human-annotated labels, and they may over-correct error-free parts, rewriting them into more fluent forms. This behavior violates the Minimum Edit Distance principle \citep{nagata2016phrase} that humans are accustomed to following when correcting grammatical errors.

Since few-shot inference is widely used to bridge alignment gaps in downstream tasks through in-context learning (ICL), LLM-based GEC systems have leveraged correction examples from databases to improve performance and interpretability \citep{davis2024prompting, song2024gee}. However, vanilla retrieval methods based on sentence embedding or k-nearest neighbors (kNN) struggle to meet the unique needs of grammatical error selection \citep{vasselli2023closer}. Grammatical errors are typically localized structural issues that are independent of word meanings, but model embeddings combine syntax and semantics into a single vector, making it fail to retrieve samples with similar error patterns. 

In this paper, we argue that despite the alignment problem in GEC tasks, language-proficient models can smoothly distinguish wrong from right and identify error patterns. This suggests that we should focus less on the generation capabilities of LLMs, but more on their internal knowledge about grammatical errors. We probe for two key questions: \textit{How does a language model encode grammatical errors internally?} and \textit{can we extract grammatical error representations that are disentangled from semantics?} 

To answer, we introduce a novel method to extract the Grammatical Error Representations (GER), a precise and interpretable representation of grammatical errors with less semantic noise, for guiding the retrieval of in-context demonstrations. Specifically, we compute error vectors (EV) by applying PCA to the difference between the hidden states of erroneous and correct tokens. We then project the hidden states of errors onto the EV to obtain the GER. As shown in \cref{fig:intro}, our GER preserves the proximity of fine-grained errors: during retrieval, each detected error aligns with similar error patterns. Additionally, over-corrected tokens are queried for similar over-correction cases in the database, improving the precision of the correction process. During inference, the number of retrieved examples dynamically adjusts based on the detected errors in the sentence, allowing for more efficient use of computational resources.

We conduct extensive experiments to demonstrate our consistent outperformance on five GEC datasets across four languages. Without additional training or generation, we obtain high-quality and interpretable demonstrations for ICL. Our results surpass state-of-the-art (SOTA) GEC retrieval methods, increasing $F_{0.5}$ by up to 9.46 points for high-resource languages like English, and by a factor of 1.20 for low-resource languages like Estonian. On open-source 8B-sized models, our approach yields results comparable to closed-source LLM baselines such as Deepseek2.5 and GPT-4o-mini, as reported by \citet{li2025explanation}.

Our contributions are summarized as follows:
\begin{itemize}
    \item We introduce a novel method to disentangle grammatical errors from semantic information and into grammatical error representations (GER), a high-quality encoding for grammatical errors.
    \item We develop an effective retriever to query examples with similar error patterns based on GER, enabling powerful ICL with LLMs across multilingual datasets.
    \item To the best of our knowledge, we are the first to explore the relationship between grammatical errors and LLM representations, offering new insights for utilizing LLMs' representations to guide GEC tasks.
\end{itemize}

\section{Related Works}

\subsection{Grammatical Error Correction}

Grammatical Error Correction (GEC) systems have wide applications in proofreading, education, and second language acquisition \citep{kaneko2022interpretability, caines2023application, liang2023chat}. Research has primarily focused on two Transformer-based approaches: sequence-to-sequence generation \citep{yuan2016grammatical, junczys-dowmunt2018approaching, li2022sequence} and sequence-to-edit tagging \citep{awasthi2019parallel, omelianchuk2020gector}. Given the local and sparse nature of grammatical errors, researchers often generate synthetic data \citep{stahlberg2024synthetic}, incorporate additional information \citep{zhang2022syngec, fei2023enhancing}, or add extra processing steps during inference \cite{lai2022type, zhou2023improving, zhang2024bidirectional, li2024detection} to boost performance. Recent work also explores LLMs for GEC, either through direct correction generation \citep{loem2023exploring} or instruction tuning \citep{fan2023grammargpt}. Despite challenges like over-correction and misalignment in LLMs \citep{vasselli2023closer}, human evaluations often rate their corrections highly \citep{zeng2024evaluating}.

\subsection{Interpretable Representations in LLMs}

Although LLMs are often seen as black boxes due to their vast number of parameters, recent research has shown that they develop emergent structures within their representations \citep{elhage2021mathematical, zou2023representation}. In the simplest case, a single dimension within the model is sufficient to characterize a specific behavior \citep{arditi2024refusal, sheng2024repeval}; more complex circuits may involve dozens of neurons distributed across different layers interacting to form meaningful components \citep{wang2023interpretability}. These interpretable components can be understood and controlled through techniques like adding, deleting, replacing, or tuning \citep{liu2024ctrla, wu2024advancing}. Our work is the first to explore and utilize LLMs' representations related to grammatical errors.

\subsection{In-Context Learning in GEC}

LLMs have demonstrated the ability to align their generated results to the knowledge domain and style of several in-context examples \citep{brown2020language, arkadiy2024iclef}. The few-shot inference paradigm avoids the additional parameters and computational costs of fine-tuning with downstream tasks. 

The selection of examples in the prompt largely affects the performance of ICL. Researchers have increased retrieval results by filtering the data, \citep{he2021efficient, peng2023semiparametric} or optimizing query encodings and retrieval algorithms \citep{li2023finding, wang2024learning}. The most helpful examples usually share similar encodings to the query, along with sufficient diversity to increase information entropy. However, for GEC tasks, the selection goal is hard to achieve. Due to the entanglement of syntax and semantics, the error encodings tend to retrieve examples with similar meanings instead of analogous error types \citep{vasselli2023closer, song2024gee}. Recent works tackle this entanglement by having models write error explanations, which are then used to retrieve errors based on the explanation embeddings \citep{li2025explanation}. Despite the improved retrieval performance, these methods still suffer from coarse sentence-level granularity and the semantic noise introduced by generated explanations. Moreover, no work has yet addressed the issue of over-correction.

\section{Methods}

In this section, we describe a novel method for extracting vectors that characterize grammatical error information and using them to create semantically neutral grammatical error representations (GER). GER from the training dataset is stored in a database, where each error is associated with its original and corrected texts. During inference, the model retrieves similar correction examples based on GER to guide corrections, with the flexibility to dynamically adjust the number of examples depending on the complexity of the input sentence. The final GEC prediction is generated by combining the retrieved examples with a correction template.

\begin{figure*}[t]
  \includegraphics[width=\linewidth]{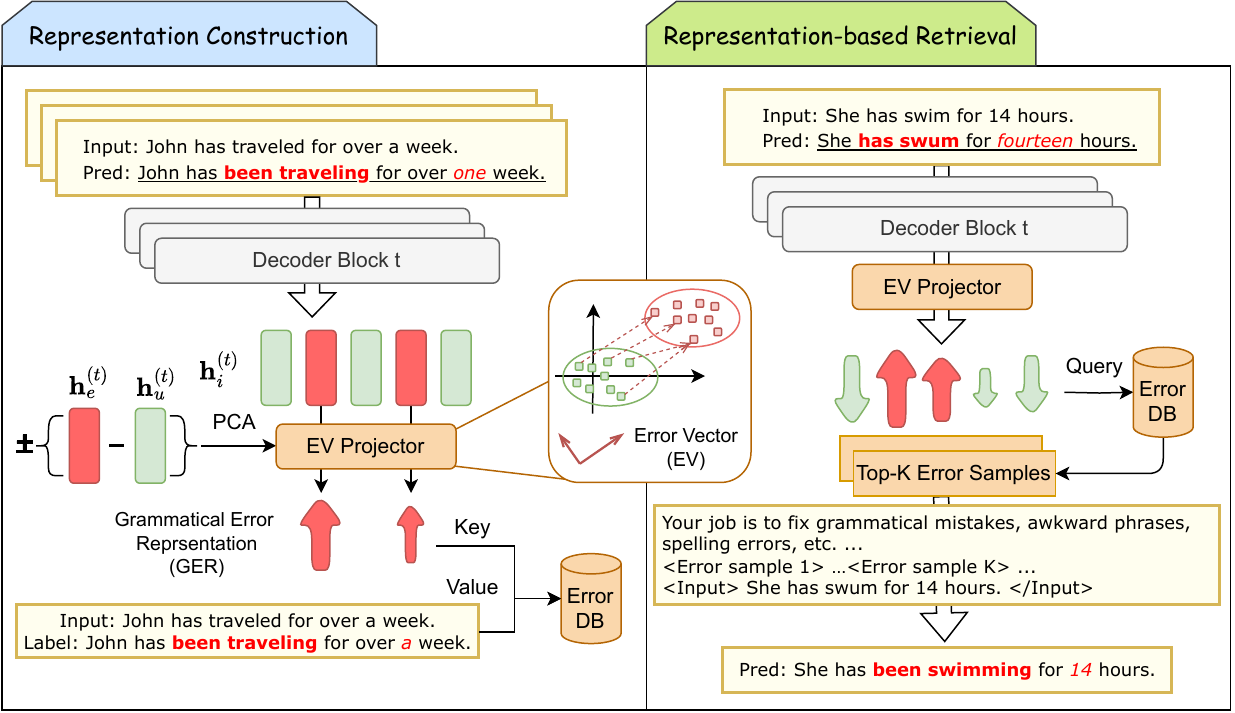}
  \caption {The pipeline for proposed representational retrieval for few-shot GEC. Left: The hidden states that best reflect the error information are extracted and transformed through PCA to obtain error vectors (EV). The projections onto EV, denoted as grammatical error representations (GER), are stored as keys in the database. Right: During inference, GER of the test input serves as the query to retrieve similar error patterns to aid correction.}
\label{fig:main}
\end{figure*}

\subsection{Extraction of Error Vectors}

Given a GEC dataset $\mathcal{S} = \{(x^{(k)}, y^{(k)})\}_{k=1}^N$, each sample consists of a potentially erroneous text $x$ and its parallel corrected text $y$. $x$ is prompted with an initial correction prompt, which can be zero-shot or filled with random initial demonstrations\footnote{The selection of examples in the initial prompt is discussed in \cref{subsec:shot}.}. During the generation of the initial prediction $\hat{y}$, we extract the hidden state at the $i$-th position from the $t$-th layer of the model, denoted as $\mathbf{h}_i^{(t)}$, obtaining the set $\mathcal{H}^{(t)}$. The choice of the specific layer $t$ is discussed in \ref{subsec:layer}. For simplicity, the subsequent formulas omit the layer index.

\begin{equation}
\label{eq:train_pred}
\hat{y} = \text{LLM}\big(\text{prompt}_\text{init}(x)\big)
\end{equation}

\begin{equation}
\label{eq:hidden}
\mathcal{H}^{(t)} = \left\{ \mathbf{h}_i^{(t)} \mid \forall i \in \{1,\dots,|\hat{y}|\} \right\}
\end{equation}

By comparing $x$ and $\hat{y}$, we identify all edits made by the LLM and collect the set of edited positions $\mathcal{E}$ and unedited positions $\mathcal{U}$. The corresponding hidden states, $\mathcal{H}_{\mathcal{E}}$ and $\mathcal{H}_{\mathcal{U}}$, contain the information necessary for the model to decide whether to correct. The difference between these sets captures the directions that guide the model from copying the original text to making corrections - precisely the information related to grammatical errors. We multiply this difference by a random sign variable $\alpha_{e,u} \in \{-1,1\}$, which randomly changes the sign to enhance the weight of the error-related directions in the principal components. 

\begin{equation}
\label{eq:pos}
\begin{aligned}
\mathcal{E} &= \left\{ i \mid \text{Align}(x, \hat{y})[i] = \text{Edited} \right\}_{i=1}^{|\hat{y}|} \\
\mathcal{U} &= \left\{ i \mid \text{Align}(x, \hat{y})[i] = \text{Unedited} \right\}_{i=1}^{|\hat{y}|}
\end{aligned}
\end{equation}

\begin{equation}
\label{eq:hidden_pos}
\begin{aligned}
\mathcal{H}_{\mathcal{E}} &= \left\{ \mathbf{h}_i \mid \forall i \in {\mathcal{E}} \right\} \\
\mathcal{H}_{\mathcal{U}} &= \left\{ \mathbf{h}_i \mid \forall i \in {\mathcal{U}} \right\}
\end{aligned}
\end{equation}

\begin{equation}
\label{eq:diff}
\Delta \mathbf{H} = \left\{ \alpha_{e,u}(\mathbf{h}_e - \mathbf{h}_u) \mid \forall e \in {\mathcal{E}}, \forall u \in {\mathcal{U}} \right\}
\end{equation}

We apply Principal Component Analysis (PCA) to the difference $\Delta \mathbf{H}$, yielding a set of principal components $\mathbf{R}$. As shown in \cref{subsec:encode}, $\mathbf{R}$ encapsulates information related to grammatical errors, with the first principal component $\mathbf{r}_1$ representing the simplicity of the error, indicating how easy it can be corrected. The first two principal components are sufficient for encoding simple error types disentangled from the text's meaning. We designate $\mathbf{R}$ as the \textbf{error vectors (EV)} of the model.

\begin{equation}
\label{eq:pca}
\Delta \mathbf{H} = \mathbf{U}\mathbf{\Sigma} \mathbf{R}^\top
\end{equation}

\subsection{Construction of GER Database}

For each correction $e \in \mathcal{E}$, we average the difference between $\mathbf{h}_e$ and all corresponding $\mathbf{h}_u \in \mathcal{H}_{\mathcal{U}}$ in the same sentence, canceling out noise from token meanings and positional embeddings. We then apply PCA, projecting onto $m$ principal components\footnote{The choice of dimensions for GER is discussed in \cref{subsec:dimension}.} to obtain the \textbf{grammatical error representation (GER)} $\mathbf{p}_e^{(m)}$. We omit dimension labeling where it is not necessary. GER serves as the key, with the corresponding pair $(x, y)$ as the label, to construct the GER database $\mathcal{D}$.

\begin{equation}
\label{eq:mean}
\Delta \mathbf{\bar{h}}_e = \frac{1}{|\mathcal{U}|} \sum_{u \in \mathcal{U}} (\mathbf{h}_e - \mathbf{h}_u)
\end{equation}

\begin{equation}
\label{eq:proj}
\mathbf{p}_e^{(m)} = \begin{bmatrix} 
\mathbf{r}_1, \mathbf{r}_2,...,\mathbf{r}_m 
\end{bmatrix}^\top \Delta \mathbf{\bar{h}}_e, \forall e \in \mathcal{E}
\end{equation}

\begin{equation}
\label{eq:database}
\mathcal{D} = \left\{ \left( \mathbf{p}_e \to (x, y) \right) \mid \forall (x,y) \in \mathcal{S}, \forall e \in \mathcal{E} \right\}
\end{equation}

\subsection{Retrieval of In-Context Demonstrations}

During inference, the test input $\widetilde{x} \in \widetilde{\mathcal{S}}$ undergoes the pipeline from \cref{eq:train_pred}-\cref{eq:diff} to obtain GER for every edit, which is then used as the query $\mathbf{q}_e$ to retrieve the $K_e$ nearest neighbors from $\mathcal{D}$.

\begin{equation}
\label{eq:knn}
\mathcal{N}(\mathbf{q}_e) = \left\{ \left( \mathbf{p}_e \to (x, y) \right)^{(j)} \right\}_{j=1}^{K_e} \subseteq \mathcal{D}
\end{equation}

Thanks to the fine-grained error encoding, we dynamically allocate the number of retrieved demonstrations $K_s$ based on the complexity of each sentence's errors. Sentences deemed error-free by the model are not assigned examples, saving computational resources for sentences with more errors. We further reveal in \cref{subsec:encode} that the magnitude of the first dimension of GER $|\mathbf{p}_e^{(1)}|$ correlates with the simplicity of the error. Therefore, we prioritize retrieval for errors that have small $|\mathbf{p}_e^{(1)}|$, further optimizing resource allocation\footnote{We describe the exact logic of dynamic selection in \cref{app:dynamic}.}.

The retrieved examples are concatenated and combined with a few-shot correction template to prompt the final GEC prediction. The inference pipeline is illustrated in \cref{fig:main}, and the prompts used are listed in \cref{app:prompt}.

\section{Experiments}

\subsection{Datasets, Models, and Metrics}
\label{subsec:dataset}

We evaluate the proposed method on five GEC datasets across four languages to test GER's ability to encode and retrieve errors. Following the multilingual setup in \citet{li2025explanation}, we process the training dataset and use LlamaIndex \citep{liu2022llamaindex} to construct the database and retriever. 

For high-resource English (EN), we use the W\&I+LOCNESS \citep{bryant2019bea} as the training dataset, and the CoNLL-14 \citep{ng2014conll} and BEA-19 \citep{bryant2019bea} datasets for testing. For medium-resource German (DE), we use the Falko-Merlin \citep{boyd2018using} dataset for both training and testing. To showcase the generalizability of our method, we also include low-resource Romanian (RO) and Estonian (ET). For Romanian, we choose the RONACC \citep{cotet2020neural} training and test datasets; for Estonian, we use the Tartu L2 learner corpus \citep{rummo2017tu} as the database and the L1 (Tartu-L1) as the test data.\footnote{The detailed statistics of GEC datasets are placed in \cref{app:dataset}.}

Since GER requires the model's internal states, all experiments are conducted using recent open-source multilingual LLMs, including Meta's Llama3.1-8B-Instruct \citep{dubey2024llama} and Tongyi's Qwen2.5-7B-Instruct \citep{yang2024qwen2}. Adhering to the dataset-specific evaluation pipeline for each language, we use the ERRANT toolkit \citep{bryant2017automatic} to align edits between initial and final predictions. For evaluation, we apply M2Scorer \citep{dahlmeier2012better} for CoNLL-14, Falko-Merlin, and Tartu-L1, while ERRANT for BEA-19 and RONACC.

Our method is compared with the following baselines:
\begin{itemize}
    \item Random: Random selection of in-context demonstrations from the database;
    \item Semantic: kNN retrieval based on input text embeddings \citep{khandelwal2020nearest};
    \item BM25: A term-based ranking function widely used in information retrieval \citep{robertson2009probabilistic};
    \item Explanation: Retrieval based on the similarity of LLM-generated explanations for erroneous sentences \citep{li2025explanation}.
\end{itemize}

All experiments are conducted in an 8-shot setting. For all baseline methods, we retrieve 4 erroneous and 4 correct examples, following \citet{li2025explanation}. Since our method dynamically determines the number of examples needed for each sentence, we retrieve 4 examples for each error and ensure that the average demonstration number is 8.

\subsection{Main Results}
\label{subsec:main}

\begin{table*}
\centering
\resizebox{\textwidth}{!}{ 
  \begin{tabular}{c|c|ccc|ccc|ccc|ccc|ccc}
\hline
\multirow{3}{*}{\textbf{Model}} & \multirow{3}{*}{\textbf{Method}} & \multicolumn{6}{c|}{\textbf{English}} & \multicolumn{3}{c|}{\textbf{German}} & \multicolumn{3}{c|}{\textbf{Romanian}} & \multicolumn{3}{c}{\textbf{Estonian}} \\
& & \multicolumn{3}{c}{\textbf{CoNLL-14}} & \multicolumn{3}{c|}{\textbf{BEA-19}} & \multicolumn{3}{c|}{\textbf{Falko-Merlin}} & \multicolumn{3}{c|}{\textbf{RONACC}} & \multicolumn{3}{c}{\textbf{Tartu-L1}} \\
\cline{3-17}
& & \textbf{P} & \textbf{R} & \textbf{F$_{0.5}$} & \textbf{P} & \textbf{R} & \textbf{F$_{0.5}$} & \textbf{P} & \textbf{R} & \textbf{F$_{0.5}$} & \textbf{P} & \textbf{R} & \textbf{F$_{0.5}$} & \textbf{P} & \textbf{R} & \textbf{F$_{0.5}$} \\
\hline
\multirow{6}{*}{\makecell{\textbf{Llama3.1}\\ \textbf{(8B)}}} & Random & 54.02 & 52.60 & 53.73 & 44.20 & 63.43 & 47.05 & 59.62 & 54.53 & 58.53 & 35.64 & 40.70 & 36.55 & 12.55 & 22.34 & 13.76 \\
& Semantic & 55.21 & 51.56 & 54.44 & 45.51 & 62.84 & 48.17 & 60.03 & 54.15 & 58.75 & 39.33 & 43.77 & 40.14 & 12.74 & 22.52* & 13.95 \\
& BM25 & 54.58 & 51.58 & 53.95 & 44.18 & 62.95 & 46.98 & 59.65 & 55.63 & 58.80 & 40.32 & 45.45 & 41.25 & - & - & - \\
& Explanation & 55.00 & 53.04 & 54.60 & 45.24 & 63.26 & 47.97 & 60.35 & 54.79 & 59.15 & 38.64 & 44.78 & 39.72 & 13.38	& \textbf{23.09} & 14.61 \\
\rowcolor{gray!30}
\cellcolor{white}
& $\text{GER-Vanilla}$ & 58.60* & \textbf{55.33} & 57.92* & 51.42* & 65.67* & 53.75* & 64.35* & 55.88* & 62.46* & 45.08* & \textbf{46.14} & 45.29* & 16.18* & 19.45 & 16.74* \\
\rowcolor{gray!30}
\cellcolor{white}
& $\text{GER-IPE}$ & \textbf{60.11} & 54.75* & \textbf{58.96} & \textbf{55.63} & \textbf{67.28} & \textbf{57.63} & \textbf{65.54} & \textbf{57.34} & \textbf{63.71} & \textbf{48.53} & 45.61* & \textbf{47.92} & \textbf{16.37} & 20.57 & \textbf{17.07} \\
\hline
\multirow{6}{*}{\makecell{\textbf{Qwen2.5}\\ \textbf{(7B)}}} & Random & 54.43 & 53.50 & 54.24 & 44.84 & 63.62 & 47.65 & 55.25 & 48.06 & 53.65 & 29.73 & 26.06 & 28.91 & 7.11 & 16.35 & 8.02 \\
& Semantic & 55.27 & 52.65 & 54.73 & 45.48 & 63.40 & 48.21 & 57.81 & 48.57 & 55.69 & 35.76 & 30.43 & 34.55 & 6.93 & \textbf{19.30} & 7.95 \\
& BM25 & 54.11 & 52.25 & 53.73 & 44.67 & 63.89* & 47.53 & 57.21 & 50.18* & 55.65 & 36.28 & 34.21* & 35.84 & - & - & - \\
& Explanation & 55.67 & 51.60 & 54.81 & 47.22 & 62.31 & 49.62 & 57.33 & 47.63 & 55.08 & 30.17 & 29.53 & 30.04 & 7.16 & 19.10* & 8.18 \\
\rowcolor{gray!30}
\cellcolor{white}
& $\text{GER-Vanilla}$ & 55.78* & \textbf{56.94} & 56.00* & 49.12* & 63.24 & 51.41* & \textbf{61.09} & 48.15 & 57.97* & 36.58* & \textbf{34.36} & 36.11* & 8.59* & 12.51 & 9.16* \\
\rowcolor{gray!30}
\cellcolor{white}
& $\text{GER-IPE}$ & \textbf{57.53} & 55.62* & \textbf{57.13} & \textbf{52.37} & \textbf{67.37} & \textbf{54.81} & 60.31* & \textbf{51.90} & \textbf{58.42} & \textbf{37.75} & 32.69 & \textbf{36.62} & \textbf{9.19} &  13.50 & \textbf{9.82} \\
\hline
\end{tabular}
}
\caption{Results on multilingual GEC datasets by different retrieval methods. "Random" refers to retrieval baseline by random selection; "Semantic", "BM25", and "Explanation" retrieve demonstrations based on text embedding, BM25 matching, and LLM-generated explanations, respectively. "$\text{GER-Vanilla}$" refers to our representation-based retrieval methods, and "$\text{GER-IPE}$" refers to GER with Initial Prompt Enhancement. The best results are marked in bold, and the second-best results are marked with an asterisk (*).}
\label{tbl:main}
\end{table*}

During preliminary experiments, we found that the construction of examples in the initial prompt significantly affects results. Thus, we present results in two configurations: "$\text{GER-Vanilla}$" refers to generating the initial predictions using the vanilla initial prompt, and "$\text{GER-IPE}$" (GER with Initial Prompt Enhancement) adds 8 randomly chosen examples into the initial prompt.

As \cref{tbl:main} demonstrates, our GER-based retrieval methods consistently outperform other baseline methods in both prompt settings. In the $\text{GER-IPE}$ setting, our method exceeds the \textbf{explanation-based} SOTA by 4.36 and 4.56 points on the English CoNLL-14 and German Falko-Merlin datasets, respectively. Moreover, the BEA-19 dataset achieves a 9.46-point higher $F_{0.5}$ than the \textit{semantic} SOTA, nearly a $20\%$ improvement. $\text{GER-Vanilla}$ still results in an improvement of around 3-5.6 points above SOTA, testifying to the effectiveness of our GER extraction and retrieval process.

On low-resource languages, GER retrieval yields even better results. For Romanian, the $F_{0.5}$ score improves by 6.67 points, while Estonian shows a 2.46 points improvement (nearly $17\%$). In $\text{GER-Vanilla}$, results are about 1 point lower but still surpass the SOTA. We hypothesize that low-resource languages benefit more from examples to help the model grasp syntax and generate corrections, as discussed in \cref{subsec:shot}.

On the Qwen2.5 model, the results follow a similar trend to Llama3.1, confirming the generalizability of our approach across models. However, the advantage is slightly lower for low-resource languages, likely due to Qwen2.5’s smaller pre-trained corpus for these languages.

\subsection{Comparison with SOTA}
\label{subsec:sota}

\begin{table}
    \centering
\resizebox{\columnwidth}{!}{ 
    \begin{tabular}{l|l|l|ccc}
\hline
\multirow{2}{*}{\textbf{Backbone}} & \multirow{2}{*}{\textbf{Method}} & \multirow{2}{*}{\textbf{Lang}} & \textbf{EN} & \textbf{DE} & \textbf{ET} \\
\cline{4-6}
& & & \multicolumn{3}{c}{\textbf{F$_{0.5}$}} \\
\hline
\multicolumn{6}{c}{\textbf{Fine-tuned GEC Single Model}} \\
\hline
gT5 xxl & \citet{rothe2021simple} & Mono & 65.7 & \textbf{76.0} & - \\
NLLB & \citet{luhtaru2024no} & Multi & 65.2 & 73.9 & \textbf{63.2} \\
BART & \citet{zhou2023improving} & Mono & \textbf{69.6} & - & - \\
\hline
\multicolumn{6}{c}{\textbf{Inference of LLMs}} \\
\hline
GPT-3.5-Turbo & \citet{davis2024prompting} & - & 57.2 & - & - \\
GPT-3.5-Turbo & \citet{tang2024ungrammatical} & - & 58.8 & - & - \\
Deepseek2.5 & \citet{li2025explanation} & - & \textbf{59.4} & 63.4 & \textbf{22.7} \\
GPT-4o-mini & \citet{li2025explanation} & - & 58.7 & \textbf{65.6} & 19.9* \\
\rowcolor{gray!30}
Llama3.1 (8B) & Ours & - & 59.0* & 63.7* & 17.1 \\
\hline
    \end{tabular}
    }
    \caption{The comparison of state-of-the-art (SOTA) models on multilingual GEC datasets. "EN", "DE", and "ET" stand for the CoNLL-14, Falko-Merlin, and Tartu-L1 datasets, respectively. Fine-tuned language models are labeled with their training data in the "Lang" column, where the "Mono" models are tuned separately for each language, and the "Multi" models with multilingual mixed data. Within each block, the best results are marked in bold, and the second-best results are marked with an asterisk (*).}
    \label{tbl:sota}
\end{table}

Current datasets reveal a persistent performance disparity in GEC tasks: while fine-tuned specialist models achieve state-of-the-art (SOTA) results across multilingual benchmarks (see \cref{tbl:sota}), in-context learning (ICL) with LLMs exhibits significant accuracy gaps. Our representational retrieval method manages to achieve results comparable to some closed-source models on high-resource English and German, including the Deepseek2.5 and GPT-4o-mini baselines reported by \citet{li2025explanation}. These promising results demonstrate the potential of utilizing interpretable components within the model to better align with human concepts and annotations of grammatical errors.

\subsection{Over-correction mitigation}

To clarify the mechanism behind our method's effectiveness, we report the True Positive (TP), False Positive (FP), and False Negative (FN) statistics from representative Llama3.1-8B runs in \cref{tbl:overcorrection}. Compared to the best-performing baseline, our GER method reduces FP by nearly $30\%$ (e.g., from 1603 to 1153 in RONACC). This indicates that the performance improvement stems primarily from substantial gains in precision, driven by a significant reduction in FP, while recall remains relatively stable (i.e., with only modest increases in TP). The mitigation of over-correction is particularly pronounced in low-resource languages such as Romanian, where models exhibit a higher propensity for overcorrecting.

\begin{table}
    \centering
\resizebox{\columnwidth}{!}{ 
    \begin{tabular}{c|ccc|ccc|ccc}
\hline
\multirow{2}{*}{\textbf{Method}} & \multicolumn{3}{c|}{\textbf{EN}} & \multicolumn{3}{c|}{\textbf{DE}} & \multicolumn{3}{c}{\textbf{RO}} \\
\cline{2-10}
& \textbf{TP($\uparrow$)} & \textbf{FP($\downarrow$)} & \textbf{FN($\downarrow$)} & \textbf{TP($\uparrow$)} & \textbf{FP($\downarrow$)} & \textbf{FN($\downarrow$)} & \textbf{TP($\uparrow$)} & \textbf{FP($\downarrow$)} & \textbf{FN($\downarrow$)}  \\
\hline
Random & 1529 & 1315 & 1389 & 3239 & 2227 & 2694 & 970 & 1752 & 1413 \\
BM25 & 1484 & 1235 & 1393 & 3311 & 2237 & 2652 & 1080 & 1603 & 1300 \\
Expl. & 1515 & 1244 & 1350 & 3258 & 2121 & 2712 & 1067 & 1694 & 1316 \\
GER & \textbf{1613} & \textbf{1098} & \textbf{1348} & \textbf{3423} & \textbf{1807} & \textbf{2540} & \textbf{1081} & \textbf{1153} & \textbf{1296} \\
\hline
    \end{tabular}
    }
    \caption{TP/FP/FN counts across datasets from representative Llama3.1-8B runs. "Expl." stands for the \textit{Explanation} baseline. For TP, the larger the better; For FP/FN, the smaller the better.}
    \label{tbl:overcorrection}
\end{table}

\subsection{Model Scalability}

To further demonstrate the effectiveness of our method on larger models, we applied GER to Qwen2.5-14B-Instruct \citep{yang2024qwen2}. The results are presented in \cref{tbl:larger}. Larger models exhibit a tendency towards excessive corrections, which can improve recall but reduce precision. By primarily mitigating over-correction, our method ensures robust performance generalization on larger models.

\begin{table}
    \centering
\resizebox{\columnwidth}{!}{ 
    \begin{tabular}{c|ccc|ccc|ccc}
\hline
\multirow{2}{*}{\textbf{Method}} & \multicolumn{3}{c|}{\textbf{EN}} & \multicolumn{3}{c|}{\textbf{DE}} & \multicolumn{3}{c}{\textbf{ET}} \\
\cline{2-10}
& \textbf{P} & \textbf{R} & \textbf{F$_{0.5}$} & \textbf{P} & \textbf{R} & \textbf{F$_{0.5}$} & \textbf{P} & \textbf{R} & \textbf{F$_{0.5}$}  \\
\hline
Random & 49.2 & 58.0 & 50.7 & 51.8 & 50.6 & 51.6 & 6.5 & 18.1 & 7.5 \\
Expl. & 50.6 & 56.2 & 51.6 & 52.9 & 52.1 & 52.7 & 6.7 & \textbf{20.3} & 7.7 \\
GER & \textbf{54.3} & \textbf{58.5} & \textbf{55.1} & \textbf{55.2} & \textbf{52.9} & \textbf{54.7} & \textbf{9.0} & 14.2 & \textbf{9.7} \\
\hline
    \end{tabular}
    }
    \caption{Results for the CoNLL-14, Falko-Merlin, and Tartu-
L1 datasets on Qwen2.5-14B. "Expl." stands for the \textit{Explanation} baseline.}
    \label{tbl:larger}
\end{table}

\section{GER Analysis}

\subsection{Encoding Capacity of GER}
\label{subsec:encode}

The different principal components calculated by PCA, referred to as error vectors (EVs), capture various levels of error-related information in natural sentences. Our preliminary exploration of the first few EVs shows that the first EV represents the model's recognition and ranking of grammatical errors, while the second EV captures simple information about error types, such as tense issues. In the following analysis section, unless stated otherwise, we use the $\text{GER-IPE}$ setup with Llama3.1-8B.

\subsubsection{The First EV: Error Detector}

\begin{figure}
    \centering
    \includegraphics[width=\columnwidth]{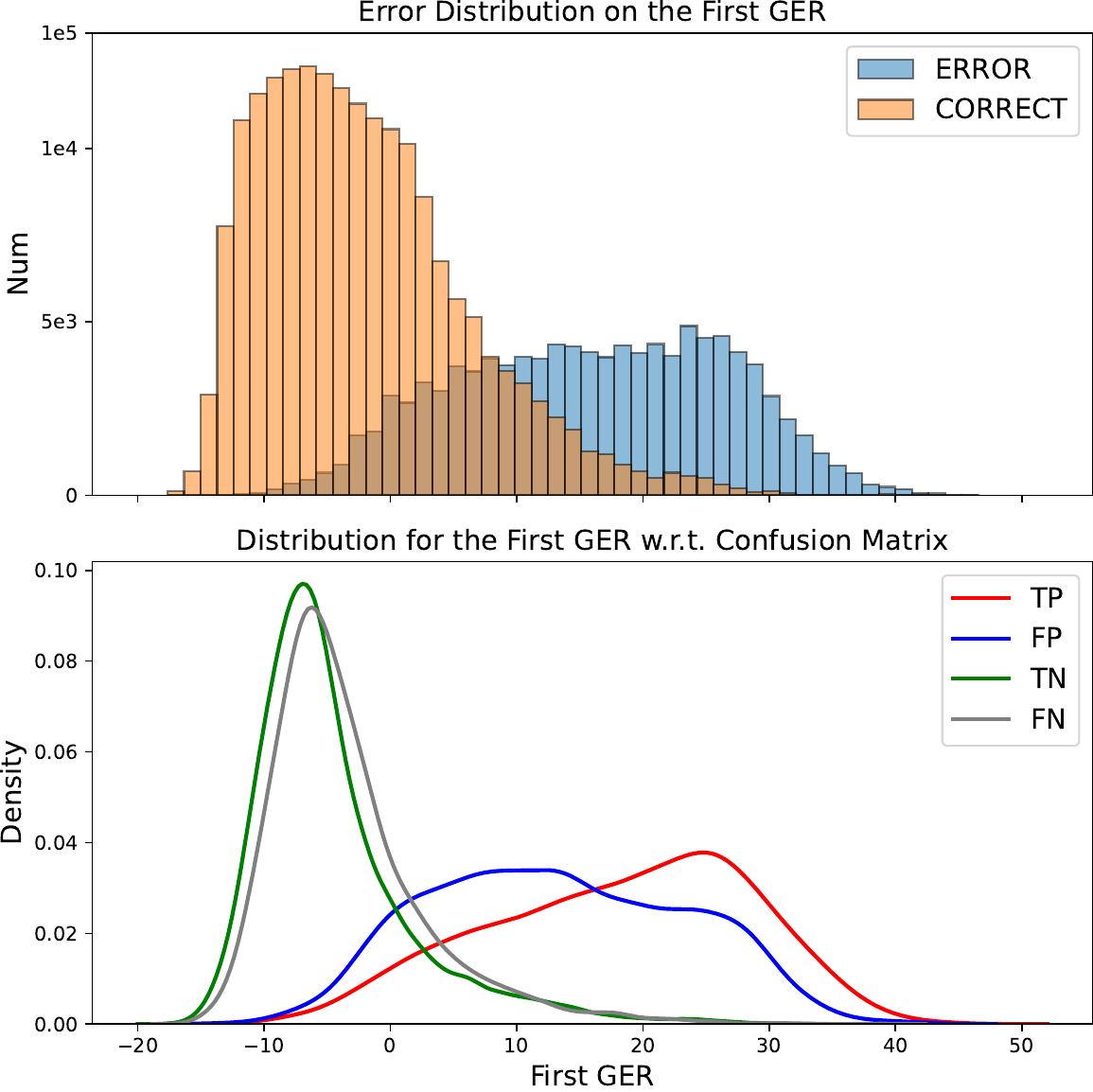}
    \caption{Distribution of the first GER component with respect to error/correct (up) and confusion matrix (down).}
    \label{fig:first_ev}
\end{figure}

We illustrate the first component of GER (first GER) obtained from the English training dataset in \cref{fig:first_ev}. The figure presents a clear boundary between erroneous and correct tokens along the direction of the first EV, achieving classification accuracy over $98\%$ for correct tokens and over $65\%$ for erroneous tokens, on par with SOTA LMs and superior to LLMs in end-to-end GED tasks \citep{luhtaru2024no}. The first GER can thus serve as an effective error detector.

Moreover, the magnitude of the first GER quantifies correction simplicity in a relatively quantitative manner. We classify predicted tokens using the confusion matrix and plot the distributions of True Positive (TP), False Positive (FP), True Negative (TN), and False Negative (FN) in \cref{fig:first_ev}. Cases with a larger first GER magnitude are more likely to represent precise corrections, whereas those with smaller values often correspond to failed corrections (FP, including over-corrections and incorrect corrections). 

Consequently, we design a dynamic demonstration selection method that prioritizes errors with small first GER values for demonstration allocation. This approach conserves computational resources for errors prone to failed corrections, which require reference to examples for successful resolution. In \cref{tbl:dynamic}, we conduct an ablation study on this selection method by comparing random example selection (Random) with prioritizing retrieval for errors having a large first GER (Reverse). The results validate the efficacy of our dynamic selection method.

\begin{table}
    \centering
\resizebox{\columnwidth}{!}{ 
    \begin{tabular}{c|ccc|ccc|ccc}
\hline
\multirow{2}{*}{\textbf{Method}} & \multicolumn{3}{c|}{\textbf{EN}} & \multicolumn{3}{c|}{\textbf{DE}} & \multicolumn{3}{c}{\textbf{ET}} \\
\cline{2-10}
& \textbf{P} & \textbf{R} & \textbf{F$_{0.5}$} & \textbf{P} & \textbf{R} & \textbf{F$_{0.5}$} & \textbf{P} & \textbf{R} & \textbf{F$_{0.5}$}  \\
\hline
Dynamic & 60.1 & \textbf{54.8} & \textbf{59.0} & \textbf{65.5} & \textbf{57.3} & \textbf{63.7} & \textbf{15.1} & \textbf{20.1} & \textbf{15.9} \\
Random & 59.8 & 52.6 & 58.2 & 64.1 & 55.5 & 62.2 & 13.9 & 20.0 & 14.8 \\
Reverse & \textbf{60.7} & 50.3 & 58.3 & 65.2 & 54.6 & 62.8 & 14.4 & 17.8 & 15.0 \\
\hline
    \end{tabular}
    }
    \caption{Ablation of different demonstration selection methods of GER.}
    \label{tbl:dynamic}
\end{table}

\subsubsection{The Second EV: Simple Error Classifier}
\label{subsec:second_ev}

\begin{figure}
    \centering
    \includegraphics[width=\columnwidth]{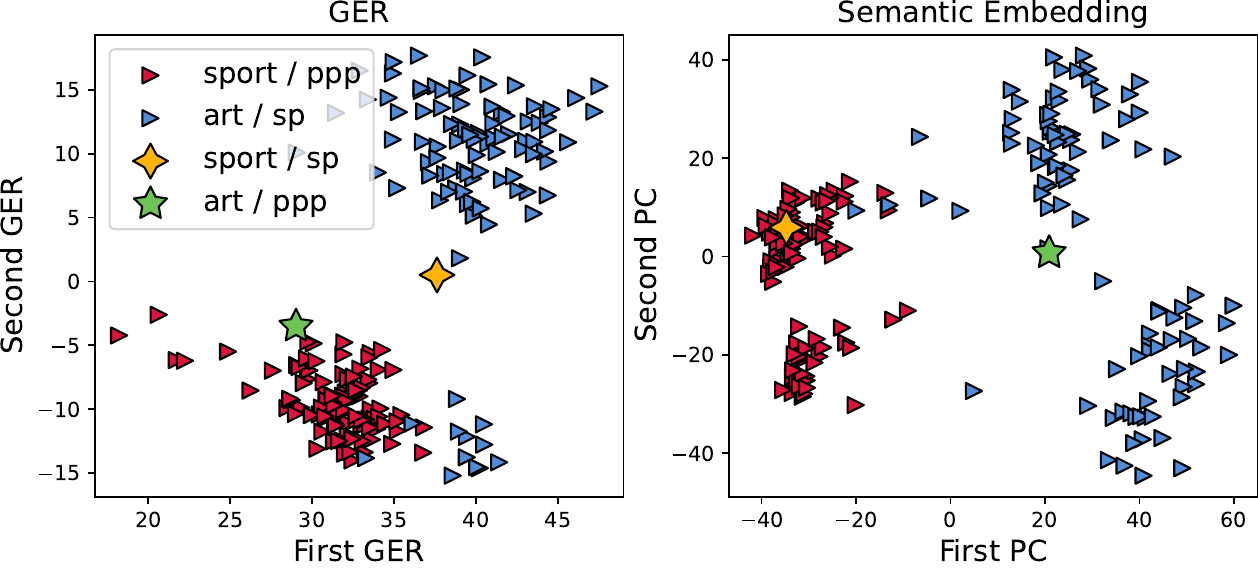}
    \caption{Distribution of different encoding methods on a manually created test set. "sport"/"art" refers to sentences in the sport/art domain, and "ppp"/"sp" refers to present perfect progressive/simple past tense errors. Cross-domain probes are marked as stars.}
    \label{fig:second_ev}
\end{figure}

On the first EV, we can distinguish between the wrong and the correct, but one dimension fails to provide detailed information. Introducing the second EV enables recognition of basic grammatical patterns. To validate this progression, we create a specialized test set\footnote{Specific samples of the test set are placed in \cref{app:case}.} containing: 
\begin{itemize}
    \item Sport-domain sentences with present perfect progressive (ppp) tense errors;
    \item Art-domain sentences with simple past (sp) tense errors.
\end{itemize}
Cross-domain probes are designed as:
\begin{itemize}
    \item Art-domain samples with ppp errors;
    \item Sport-domain samples with sp errors.
\end{itemize}
\cref{fig:second_ev} shows that while semantic embeddings retrieve semantically similar but error-mismatched examples, our 2-dimensional GER successfully clusters analogous errors across domains, demonstrating the proximity and semantic neutrality of GER.

\subsubsection{Dimensionality Trade-offs in GER}
\label{subsec:dimension}

\begin{table}
    \centering
\resizebox{\columnwidth}{!}{ 
    \begin{tabular}{c|ccc|ccc|ccc}
\hline
\multirow{2}{*}{\textbf{Dim.}} & \multicolumn{3}{c|}{\textbf{EN}} & \multicolumn{3}{c|}{\textbf{DE}} & \multicolumn{3}{c}{\textbf{ET}} \\
\cline{2-10}
& \textbf{P} & \textbf{R} & \textbf{F$_{0.5}$} & \textbf{P} & \textbf{R} & \textbf{F$_{0.5}$} & \textbf{P} & \textbf{R} & \textbf{F$_{0.5}$}  \\
\hline
128 & 59.5 & 54.5 & 58.4 & 65.2 & 57.3 & 63.4 & 14.4 & 19.4 & 15.2 \\
256 & 59.7 & 53.6 & 58.4 & 65.2 & 57.2 & 63.4 & \textbf{15.1} & 20.1 & \textbf{15.9} \\
512 & 59.8 & 54.3 & 58.6 & \textbf{65.5} & 57.3 & \textbf{63.7} & 14.7 & 20.1 & 15.5 \\
1024 & \textbf{60.1} & \textbf{54.8} & \textbf{59.0} & 65.4 & \textbf{57.4} & 63.6 & 14.9 & 20.4 & 15.8 \\
2048 & 60.0 & 54.4 & 58.8 & 65.1 & 56.9 & 63.3 &  14.3 & \textbf{20.7} & 15.2 \\
\hline
    \end{tabular}
    }
    \caption{Results across different dimensional configurations of GER.}
    \label{tbl:dimension}
\end{table}

Increasing the dimensionality of GER ($m$ in $\mathbf{p}_e^{(m)}$) enhances its ability to encode fine-grained error patterns, but simultaneously amplifies the semantic noise it contains, causing GER to extract examples with semantic similarities over those sharing similar error types. Experimental results across different dimensional configurations are presented in \cref{tbl:dimension}: the more resources the model has about a particular language, the more dimensions it needs to encode errors in that language. At reduced dimensions, GER fails to distinguish complex errors; on the other hand, when the dimensions are too large, GER can identify some nuanced error cases but introduce more error-irrelevant samples, resulting in higher recall and lower precision.

\subsection{Layer Selection}
\label{subsec:layer}

\begin{figure*}[t]
  \includegraphics[width=\linewidth]{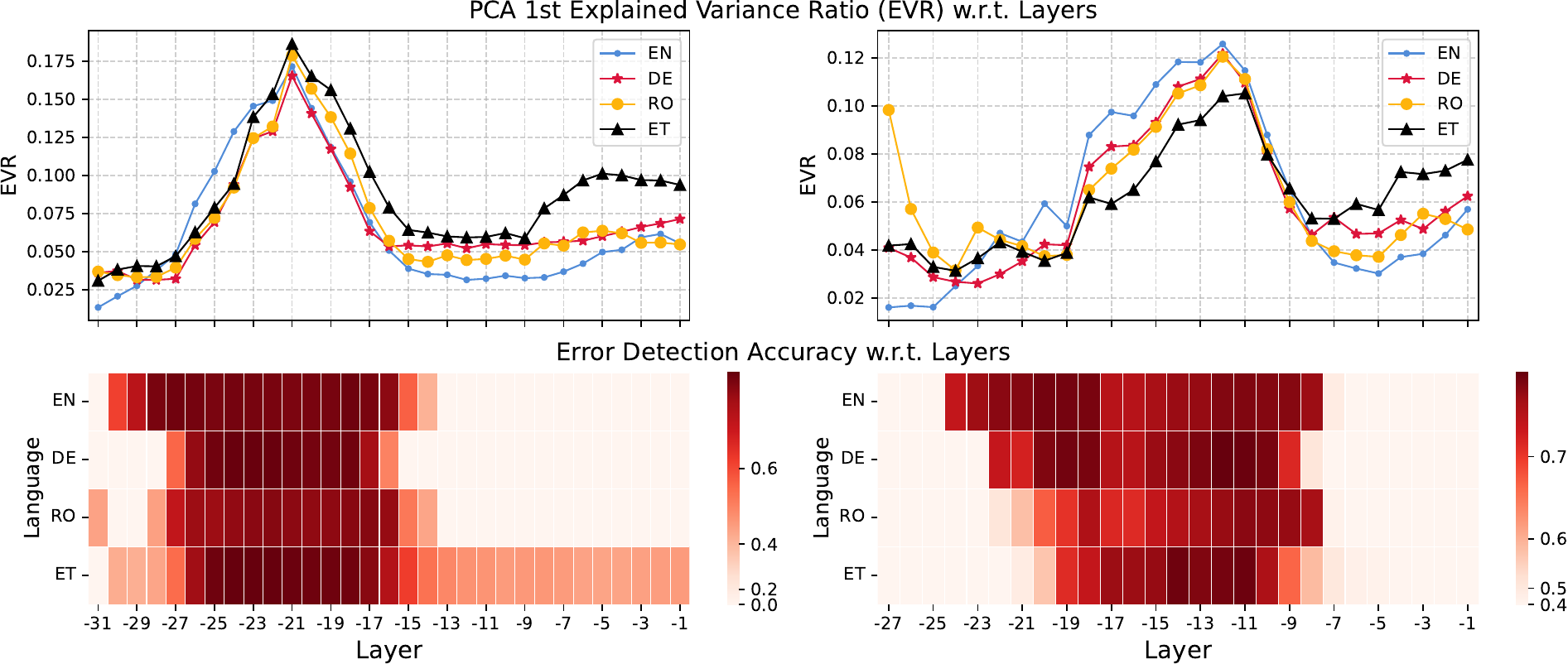}
  \caption{Upper: The explained variance ratio of the first principal component in PCA (first EVR) for layers. Lower: Accuracy of grammatical error detection task in each layer. We observe similar patterns for the trend of first EVR and error detection accuracy in Llama3.1 (left) and Qwen2.5 (right).}
  \label{fig:layer}
\end{figure*}

We select the layer used to extract GER based on the performance of grammatical error detection. The error detection performance with respect to each layer of the model is juxtaposed with the explained variance ratio of the first principal component in PCA (first EVR) in \cref{fig:layer}. From the upper figures, a spike of the first EVR is clearly depicted, coinciding with the most accurate layer in the lower images. The specific choice of layer differs with each model but remains highly consistent across languages within the same model, and the selected layers are in the middle of each model (the 21st layer for 32-layer Llama3.1, and the 12th layer for 28-layer Qwen2.5). This suggests to us that there are specific components within the layer that are responsible for understanding and processing grammatical error information. We leave further research to future work.

\subsection{Demonstration Selection for Initial Prompt}
\label{subsec:shot}

\begin{figure}
    \centering
    \includegraphics[width=\columnwidth]{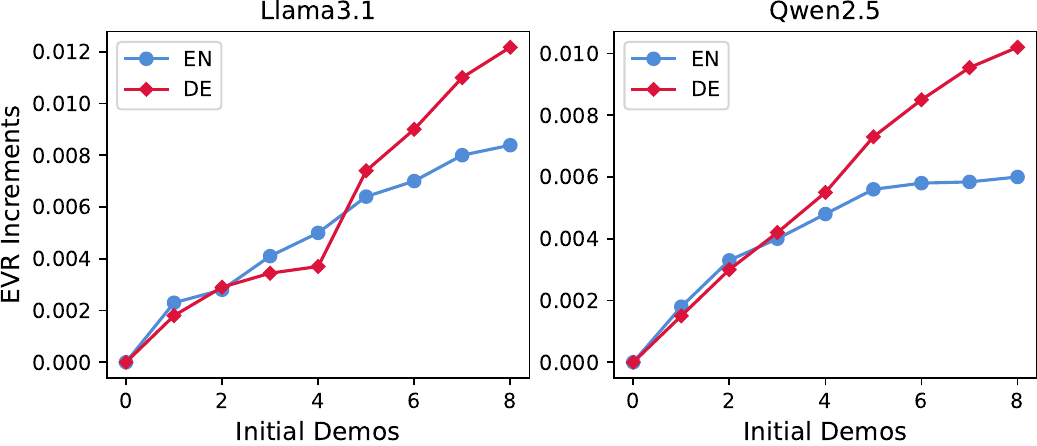}
    \caption{EVR increments of n-shot initial demonstrations relative to 0-shot.}
    \label{fig:shot}
\end{figure}

As observed in \cref{subsec:main}, even randomly selected examples in the initial prompt significantly improve results, although they affect the initial prediction and not the final output. We attribute this improvement to two factors: first, the few-shot initial prompt helps activate the model's correction capability and aligns the generated outputs with the example format. This alignment is particularly noticeable in low-resource languages such as Estonian, where zero-shot predictions usually include English tokens, introducing noise that hinders the PCA process for extracting EV. Second, from within the model, the initial prompt aligns EV inside the model toward the actual error space. \cref{fig:shot} reveals that the first explained variance ratio (EVR) increases as more initial examples are added, indicating that the model is refining its error space with each new demonstration. This suggests that the examples selected by GER may help the model better characterize the error space, which can be used iteratively in another round of generation to optimize EV. We leave this iterative approach for future work.

\section{Conclusion}

In this paper, we delve into the internals of LLMs and develop a novel method for extracting precise and interpretable grammatical error representations (GER) with less semantic noise. The effectiveness of GER in encoding fine-grained error patterns enables the retrieval of high-quality error demonstrations, improving the few-shot performance of LLMs on GEC across diverse language settings. 

Our preliminary exploration and successful utilization of LLMs' internal states highlight the potential of utilizing the model’s inherent knowledge to strengthen GEC performance, alignment, and interpretability, all without the need for additional components or training resources.

\section*{Limitations}

Our work explores and leverages the knowledge related to error correction within large models. However, the few-shot GEC capabilities of LLMs are far from fully realized. The latter dimensions of our proposed error vectors contain detailed, fine-grained knowledge about error classification and correction, but they are difficult to separate, visualize, and utilize effectively. In addition, we did not address the scenario where long sentences with multiple errors outpace the utility of the 8-shot examples. In such cases, slicing the long sentence into smaller segments may yield better performance. 

While we have encoded errors and used them for example retrieval in this work, the error information could be applied more broadly in the model's prediction pipeline, such as in controlling the decoding process. Future work could investigate simpler ways of representing error information, or develop methods to comprehensively combine and summarize this information for more effective manipulation of model-generated grammatical error corrections.

\section*{Acknowledgments}

This work was supported by National Natural Science Foundation of China (62036001) and National Science and Technology Major Project (No. 2022ZD0116308) . The corresponding author is Houfeng Wang.

\bibliography{custom}

\appendix

\section{Experimental Settings}

\subsection{Dataset Statistics}
\label{app:dataset}

Our dataset usage is shown in \cref{tbl:dataset}. The training data samples used to construct the database are initially filtered by length with a minimum of 10 to ensure quality. 

\begin{table*}
    \centering
    \begin{tabular}{l|lll|ll}
\hline
 & \multicolumn{3}{c|}{\textbf{Training Dataset (As Database)}} & \multicolumn{2}{c}{\textbf{Test Dataset}} \\
\hline
\textbf{Language} & \textbf{Name} & \textbf{$\#$Erroneous} & \textbf{$\#$Correct} & \textbf{Name} & \textbf{$\#$Total} \\
\hline
\multirow{2}{*}{\textbf{English}} & \multirow{2}{*}{W\&I+LOCNESS} & \multirow{2}{*}{20185} & \multirow{2}{*}{6839} & CoNLL-14 & 1312 \\
 & & & & BEA-19 & 4477 \\
\textbf{German} & Falko-Merlin & 11801 & 1916 & Falko-Merlin & 2337 \\
\textbf{Romanian} & RONACC & 6974 & 108 & RONACC & 1519 \\
\textbf{Estonian} & Tartu-L2-Corpus & 7156 & 4 & Tartu-L1-Corpus & 1453 \\
\hline
    \end{tabular}
\caption{The statistics of GEC dataset used in experiments. For the training datasets, $\#$Erroneous represents the number of erroneous samples, and $\#$Correct refers to the number of correct samples. For the test datasets, $\#$Total indicates the total number of samples.}
\label{tbl:dataset}
\end{table*}

\subsection{Language Diversity}
\label{app:diversity}

Our language selection aligns with prior multilingual GEC studies \citep{luhtaru2024no, stahlberg2024synthetic}, taking into account the diversity of language families.
\begin{itemize}
    \item Germanic (English, German) and Romance (Romanian) languages: Both Indo-European, but from different branches.
    \item Uralic (Estonian): a non-Indo-European language with agglutinative grammar and no grammatical gender, unlike the others. As a linguistically distant and low-resource language, Estonian showcases the breadth of GER’s applicability.
\end{itemize}
We acknowledge the value of testing additional languages (e.g., Czech, Chinese) and will explore this in future work.

\subsection{Model Settings}
\label{app:model}

We utilize open-source LLMs such as \href{https://huggingface.co/meta-llama/Llama3.1-8B-Instruct}{Llama3.1-8B-Instruct} and \href{https://huggingface.co/Qwen/Qwen2.5-7B-Instruct}{Qwen2.5-7B-Instruct} to implement representation extraction and demonstration retrieval. 

To ensure reproducibility, we applied deterministic decoding (with temperature set to 0 and top\_p set to 1.0) during inference. For the "Random" baseline, samples were selected using three different random seeds, and the results were averaged.

\subsection{Prompt Settings}
\label{app:prompt}

\begin{table*}
\centering
\resizebox{\linewidth}{!}{ 
    \begin{tabular}{l}
You are a language expert who is responsible for grammatical, lexical, and orthographic error corrections \\
given an input sentence. Your job is to fix grammatical mistakes, awkward phrases, spelling errors, etc. \\
following standard written usage conventions, but your corrections must be conservative. \\
Please keep the original sentence (words, phrases, and structure) as much as possible. \\
The ultimate goal of this task is to make the given sentence sound natural to native speakers \\
without making unnecessary changes. Corrections are not required when the sentence is already \\
grammatical and sounds natural. \\
There is an erroneous sentence between '<erroneous sentence>' and '</erroneous sentence>'. \\
Then grammatical errors in the erroneous sentence will be corrected. \\
The corrected version will be between '<corrected sentence>' and '</corrected sentence>'. \\
<erroneous sentence> {text}</erroneous sentence> \\
<corrected sentence> {label}</corrected sentence> \\
... \\
<erroneous sentence> {text}</erroneous sentence> \\
<corrected sentence> {label}</corrected sentence> \\
<erroneous sentence> {source}</erroneous sentence> \\
<corrected sentence> \\
    \end{tabular}
}
\caption{The prompts for the proposed method. \{text\} and \{label\} denote the input text and correct sentence (label) for labeled GEC data. \{source\} represents the test input text.}
\label{tbl:prompt}
\end{table*}

Throughout the entire experiment pipeline, we use the same prompt for GEC task as prior works \citep{tang2024ungrammatical, davis2024prompting, li2025explanation}, to form a fair comparison. The correction prompt is shown in \cref{tbl:prompt}.

\subsection{Dynamic Selection Setting}
\label{app:dynamic}

Dynamic example selection was introduced to ensure fair benchmarking against prior 8-shot baselines. During inference:
\begin{itemize}
    \item Given a test set of size $N$ and $K_e$ retrieved samples per edit, we obtain the GER for each edit in the test set and sort them in ascending order based on the first dimension of GER.
    \item Then, we select the top $N*K/K_e$ edits and use their corresponding samples to extract demonstrations.
\end{itemize}

\section{Time Efficiency}
\label{app:efficiency}

Our GER method can be divided into two parts:
\begin{itemize}
    \item Example Selection: Requires one forward pass over test data to extract GER. Compared to previous methods (e.g., \citet{li2025explanation}), which need to generate explicit explanations, our approach achieves a 50x speedup (average explanation length $ L\approx50$ in \citet{li2025explanation}).
    \item Few-shot Inference: With selected demonstrations, our inference latency matches that of standard 8-shot inference, without additional overhead.
\end{itemize}

\section{Cross-domain demonstration set}
\label{app:case}

\begin{table*}
\centering
\resizebox{\linewidth}{!}{ 
    \begin{tabular}{c|c|c}
    \hline
    \textbf{Domain} & \textbf{Error Type} & \textbf{Case} \\
    \hline
    \multirow{2}{*}{\textbf{Sport}} & \textbf{ppp} & \makecell[l]{Input: I have jogged along the riverbank for 45 minutes. \\Label: I \textcolor{red}{have been jogging} along the riverbank for 45 minutes.} \\
    \cline{2-3}
    & \textbf{sp} & \makecell[l]{Input: Yesterday, she try to hold her breath underwater. \\Label: Yesterday, she \textcolor{red}{tried} to hold her breath underwater.}\\
    \hline
    \multirow{2}{*}{\textbf{Art}} & \textbf{ppp} & \makecell[l]{Input: Marcel Duchamp submits a urinal to an art show in 1917. \\Label: Marcel Duchamp \textcolor{red}{submitted} a urinal to an art show in 1917.}\\
    \cline{2-3}
    & \textbf{sp} & \makecell[l]{Input: For the entire week, Georgia O'Keeffe has painted her first giant flower close-up. \\Label: For the entire week, Georgia O'Keeffe \textcolor{red}{has been painting} her first giant flower close-up.}\\
    \hline
    \end{tabular}
}
\caption{Examples from the manually constructed test set used in \cref{subsec:second_ev}.}
\label{tbl:domain}
\end{table*}

In \cref{subsec:second_ev}, we used the web version of \href{https://chat.deepseek.com}{Deepseek-v3} to build 100 sport-domain sentences with present perfect progressive (ppp) tense errors, and 100 art-domain sentences with simple past (sp) tense errors. We then created cross-domain probes such as art-domain samples with ppp errors and sport-domain samples with sp errors to show the proximity and semantic neutrality of our GER. The created cases are demonstrated in \cref{tbl:domain}.

\end{document}